\documentclass{article} 
\usepackage{times}
\usepackage{ron_refs}
\usepackage{graphicx}
\usepackage{float}
\usepackage[hidelinks]{hyperref}
\usepackage{url}
\usepackage{subcaption}
\usepackage{amsmath} 
\usepackage{placeins}
\usepackage{tabularx}
\usepackage{wrapfig}

\title{ImageNet-trained deep neural network exhibits illusion-like response to the Scintillating Grid}

\author{Eric Sun \\
	Department of Physics\\
	Department of Chemistry and Chemical Biology\\
	Harvard University\\
	Cambridge, MA 02138, USA \\
	\texttt{eric\_sun@college.harvard.edu} \\
	\And
	Ron Dekel \\
	Department of Neurobiology \\
	Weizmann Institute of Science \\
	Rehovot, PA 7610001, Israel \\
	\texttt{ron.dekel@weizmann.ac.il} \\
}

\begin{document}
	\maketitle
    

	\begin{abstract}
    Deep neural network (DNN) models for computer vision are now capable of human-level object recognition. Consequently, similarities in the performance and vulnerabilities of DNN and human vision are of great interest. Here we characterize the response of the VGG-19 DNN to images of the Scintillating Grid visual illusion, in which white dots are perceived to be partially black. We observed a significant deviation from the expected monotonic relation between VGG-19 representational dissimilarity and dot whiteness in the Scintillating Grid. That is, a linear increase in dot whiteness leads to a non-linear increase and then, remarkably, a decrease (non-monotonicity) in representational dissimilarity. In control images, mostly monotonic relations between representational dissimilarity and dot whiteness were observed. Furthermore, the dot whiteness level corresponding to the maximal representational dissimilarity (i.e. onset of non-monotonic dissimilarity) matched closely with that corresponding to the onset of illusion perception in human observers. As such, the non-monotonic response in the DNN is a potential model correlate for human illusion perception. 

	\end{abstract}


\section{Introduction}
    
Deep neural network (DNN) models are capable of besting human champions in chess \cite{schmidhuber_deep_2015} and Go \cite{silver_mastering_2016} and reaching superhuman levels of accuracy in image classification and object recognition tasks \cite{ciregan_multi-column_2012, he_delving_2015, eagleman_visual_2001, schmidhuber_deep_2015}. The comparable performances of DNNs and humans are reflected by several similarities in their computational architecture such as having isolated computational units or neurons organized hierarchically into layers. Possibly as a result of these similarities, and by virtue of the similarity in object recognition accuracy as compared to humans \cite{szegedy_going_2015, he_deep_2016}, DNNs have been proposed as models for several aspects of human vision including shape recognition \cite{kubilius_deep_2016} and visual perceptual learning \cite{wenliang_deep_2018} among others \cite{yamins_using_2016,turner_stimulus-_2019}. Several studies have suggested correlations between DNN computation stages and neural activity in primate visual areas in processing the same images \cite{cadieu_deep_2014, yamins_performance-optimized_2014, guclu_deep_2015, martin_cichy_dynamics_2017}. These results motivate the search for other visual intersections, and possibly vulnerabilities, that are shared between human and machine models. 
 
Human perception can exhibit large deviations from what is considered to be physical reality; these deviations are often referred to as visual illusions \cite{gregory_putting_1991}. The study of such illusions may provide insight into the constraints and mechanisms of human visual processing \cite{eagleman_visual_2001}. In a similar manner, DNNs are also prone to ``illusions", which include seemingly unrecognizable images that have been generated using adversarial methods to mislead DNN image classifiers \cite{szegedy_intriguing_2013, nguyen_deep_2015} and natural images that exploit flaws in current classifiers \cite{hendrycks_natural_2019}. Further, some images generated to mislead a DNN have also led to mistaken classification by time-limited human observers
\cite{elsayed_adversarial_2018}. However, to our knowledge, there has yet to be an examination of how the representation of images within the DNN model differs for images that clearly exhibit illusion perception in humans as compared to non-illusion images. Here we explore the Scintillating Grid, a human visual illusion in that regard.

The Scintillating Grid illusion (Fig.~\ref{fig:fig1}a) induces an illusory perception of scintillating black dots within white grid dots \cite{schrauf_scintillating_1997}. The Scintillating Grid is a stronger variant of the famous Hermann Grid, which exhibits a similar effect at the intersections of grid lines \cite{spillmann_hermann_1994}. The Hermann Grid is structurally identical to the Scintillating Grid with the exception of dots at the intersections. 
    
In this study, we characterize a potential model correlate for human illusion perception in the VGG-19 DNN. Using a setup where images with increasing whiteness of a masked region are compared to a standard image, we analyze the VGG-19 representation of Scintillating Grid illusion and control images and discover an illusion-specific deviation from the monotonic relationship that is expected from linear increases in pixel difference (i.e.~whitening). We introduce additional control setups, compare the deviation to the illusion effect in human perception, and examine the propagation of the deviation across the VGG-19 network architecture. Our findings suggest several similarities between VGG-19 and human responses to the Scintillating Grid and potentially offer a fresh perspective regarding the origin of the Scintillating Grid illusion effect in visual systems.


\section{Methods}

\subsection{VGG-19 DNN}
We chose to examine representations of VGG-19, which has been suggested as a model correlate for human categorization of competing images \cite{gruber_perceptual_2018}. VGG-19 is a DNN with 19 layers consisting of a stack of convolutional layers followed by three fully connected layers and a final soft-max layer \cite{simonyan_very_2014}. The VGG models produced top performances in both localization and classification tracks at the 2014 ImageNet Challenge. We utilized the standard VGG-19 model accessed through Matlab (MatConvNet \cite{vedaldi_matconvnet:_2015}) that was pre-trained on the $\sim$1.3 million images of 1,000 image classes of ImageNet \cite{deng_imagenet:_2009}. Most analyses were performed on the output representation of the final fully-connected layer (\texttt{fc8}), since it was the closest layer to the network output, and hence presumably the most similar to visual perception. All image stimuli used were compressed to dimensions of $224\times 224$ pixels.

\subsection{Dot whiteness experimental setup}
The grid illusion images were generated at a size of $768 \times 768$ pixels and then re-sized to $224 \times 224$ pixels to conform to VGG-19 input requirements. Here we denote pixel whiteness with the symbol $\gamma$. The Scintillating Grid stimulus was set to default parameters:
\begin{itemize}
\item Dots: 25 dots organized in a 5x5 grid, each of diameter 30 pixels (prior to downsizing) and whiteness $\gamma=1.0$ (white); each with a concentric border of width 1 pixel (prior to downsizing) and whiteness $\gamma=0.8$ to prevent shape loss when dot whiteness matches line whiteness.
\item Lines: 10 lines organized to intersect at dot positions, each of width 15 pixels and whiteness $\gamma=0.5$ (gray).
\item Background: whiteness $\gamma=0.0$ (black).
\end{itemize}
The dot elements were masked and their whiteness was varied along 21 uniform $\gamma$ values between black ($\gamma=0.00$) and white ($\gamma=1.00$) with $\Delta \gamma = 0.05$. These images were compared to the reference image with black dots ($\gamma=0.00$) and the $L_1$ distance in the VGG-19 representations (referred to as the representational dissimilarity, $R$) was measured (Fig.~\ref{fig:fig1}b).

To maintain contrast boundaries between dots and lines even when whiteness of dots and lines was the same, we introduced a one-pixel border of whiteness $\gamma=0.8$ around all dots prior to image downsizing. This manipulation preserved illusion perception in humans as evident by inspection (see Fig.~\ref{fig:fig1}a). We conclude that the observed deviation in the representational dissimilarity $R$ is not significantly influences by shape loss, since the peak of $R$ occurred at $\gamma=0.55$ and not $\gamma=0.8$ (i.e.~border whiteness).

For the natural and synthetic images, we selected a masked region consisting of 5-20 percent of pixels that were approximately white ($\gamma \approx 1$) using a heuristic threshold on the pixel value. The whiteness of the masked region was varied along 21 intervals to reflect the grid illusion setup. Similarly, each whiteness variant of the original image was compared to the same image with a black ($\gamma=0$) masked region to obtain a representational dissimilarity $R$ measure (see Section 2.4).

\subsection{Image stimuli sets}
We used a set of 30 illusion images and a set of 30 control images. The illusion stimuli set consisted of diverse Scintillating Grid variants including grids with translation, increased dot size, altered background color, different scales, and different dot array dimensions. The control stimuli set included 19 natural and synthetic images and 11 illusion controls (i.e.~grid images where no illusory perception was present for human observers). Natural and synthetic images included animals, humans, plants, and also randomly generated square grids or checkerboard patterns. The natural images were selected by an independent party from non-ImageNet sources. The illusion control images consisted primarily of Scintillating Grid variants with the grid lines removed. Representative images from each of the stimuli sets are available in the Supplementary and the full stimuli set is included in the public repository: \url{https://github.com/sunericd/dnn-illusion}.

\subsection{Representational Dissimilarity}
To quantify the dissimilarity between VGG-19 \texttt{fc8} representations of two images, we used the $L_1$ distance. This distance was referred to as the representational dissimilarity $R$. Specifically, the metric was calculated as the mean absolute difference of the neuron outputs $a_{ijk}$ and $b_{ijk}$ of the two images $A$ and $B$, given the $M \times N \times K$ neurons in the layer (rows, columns, convolution kernels).
\begin{equation}\label{R}
    R = \frac{\sum\limits_{i=1}^M \sum\limits_{j=1}^N \sum\limits_{k=1}^K |a_{ijk} - b_{ijk}|}{MNK}
\end{equation}

\subsection{Deviation magnitude and area}
\begin{figure}
    \centering
    \includegraphics[width=\textwidth]{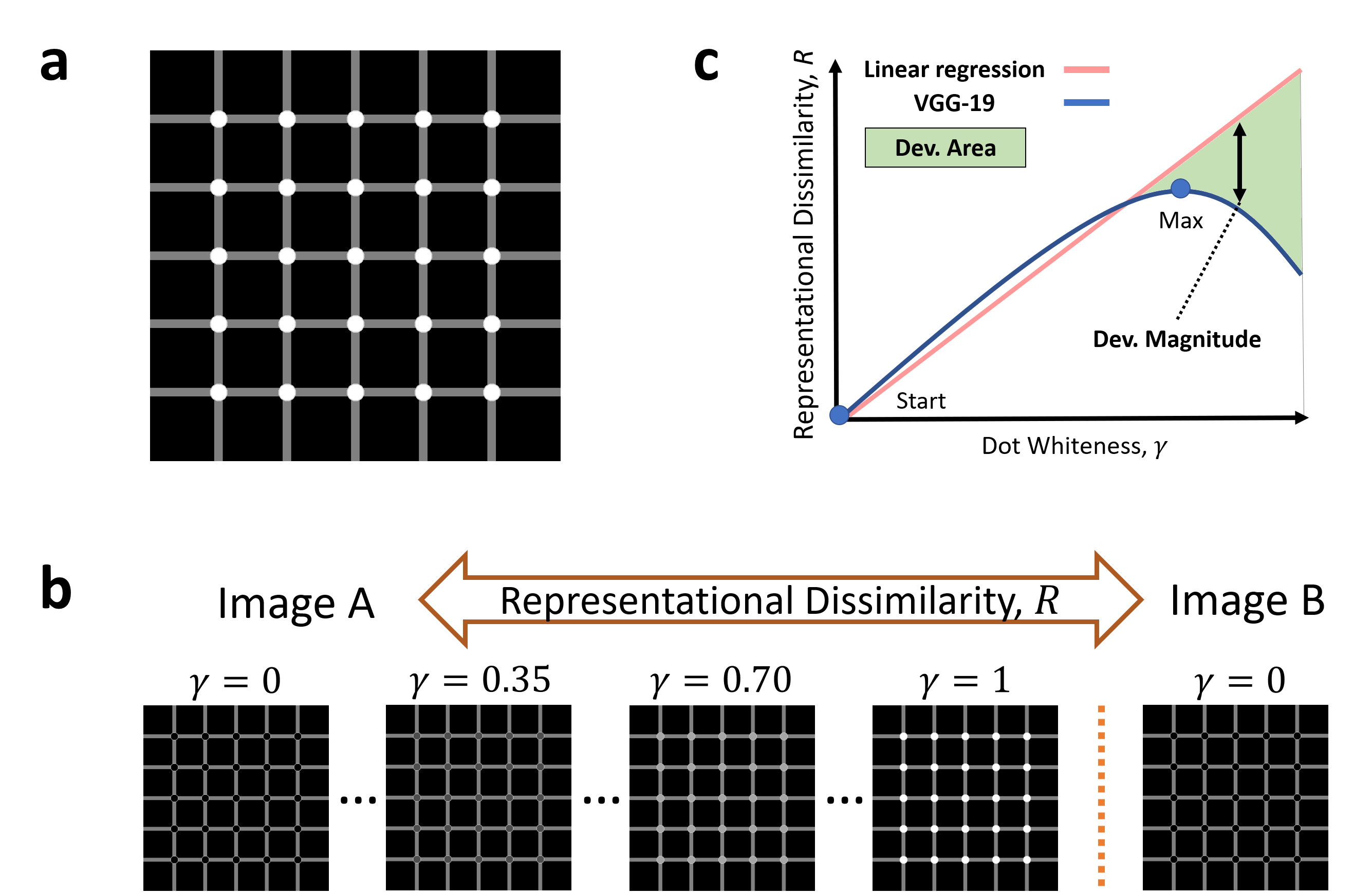}
    \caption{Scintillating Grid and experimental protocols. (a) The Scintillating Grid visual illusion exhibits illusory scintillation of black dots within the white grid dots. (b) Schematic representation of the experimental setup. Representational dissimilarity, denoted $R$, was calculated as the $L_1$ distance of the VGG-19 representation (layer \texttt{fc8}) between two images. One image had a masked region of varying whiteness (from $\gamma=0.00$ through $\gamma=1.00$ by $\Delta \gamma=0.05$) and the other image was constant with a black masked region ($\gamma=0.00$ throughout). For the Scintillating Grid (as in illustration), the masked regions were the grid dots. (c) Schematic representation of deviation magnitude $d(\gamma)$ and deviation area $D$ measurements. $d(\gamma)$ was measured as the distance between the $R$ value of a linear regression on the $(\gamma, R)$ values up to $\gamma_{\text{max}}$ and the VGG-19 $R(\gamma)$ for the given $\gamma$. $D$ was measured as the area between the linear regression and the VGG-19 dissimilarity curve and represented the accumulated magnitude of the deviation for all $\gamma$ less than the $\gamma$ at maximum $R$.}
    \label{fig:fig1}
\end{figure}

To quantify deviations from the expected relation between dot whiteness and representational dissimilarity $R$ with respect to a blackened image, we made two assumptions: 1) In the absence of illusion-like deviation, $R$ increases approximately linearly with increased whiteness, and 2) an illusion-like deviation contributes to a depressed $R$ since the perception of black dots enforces greater similarity to the black ($\gamma=0$) dot grid (i.e. illusory white dots are more similar than regular white dots to black dots). An approximation of the non-illusory representational dissimilarity as a function of dot whiteness was obtained using a linear regression of representational dissimilarity values from the initial dot whiteness ($\gamma=0$) through the dot whiteness at maximal $R$ ($\gamma_M$ such that $R(\gamma_M)=\max(R)$). We assumed linearity in this range of $\gamma$ values since there was effectively no illusory perception for human observers at low dot whiteness values \cite{sun_characterizing_2019}. The deviation from this expected linear relationship was measured by subtracting the observed representational dissimilarity $R_{\text{observed}}$ in VGG-19 from the expected linear regression $R_{\text{linear}}$ (Fig.~\ref{fig:fig1}c):

\begin{equation}\label{d}
    d(\gamma) = R_{\text{linear}}(\gamma) - R_{\text{observed}}(\gamma)
\end{equation}

We refer to this difference between the linear and observed $R$ values as the {\it deviation magnitude} $d(\gamma)$. Similarly, we refer to the positive area of the $d(\gamma)$ curve as the {\it deviation area} $D$. The deviation area represents the magnitude and depth (range of whiteness intervals) for the illusion-like deviation. $D$ was calculated as the integral of the deviation magnitude $d$ from the point of maximum representational dissimilarity $\gamma=\gamma_{\text{maxR}}$ to the final dot whiteness $\gamma=1$ (Fig.~\ref{fig:fig1}c):

 \begin{equation}\label{D}
     D = \int_{\gamma=\gamma_{\text{maxR}}}^{\gamma=1} d(\gamma) \quad d\gamma
 \end{equation}
We approximated this integral using numerical integration with trapezoidal quadrature.

\subsection{VGG-19 stage and layer analysis}
To examine the extent of the deviation in representational dissimilarity $R$ across the architecture of VGG-19, we compared the deviations at different layers, and at different neurons within a layer. Direct layer comparisons were achieved by scoring the $R$ deviation of each layer with the deviation area $D$. The $D$ was then normalized with respect to the representational dissimilarity of the final white dot grid ($\gamma=1$) for each layer respectively. This normalization was done to adjust for differences in the magnitudes of activation between layers. Similarly, comparisons at the neuronal level were achieved by using the deviation area normalized by the final ($\gamma=1.0$) $R$ value for each neuron. In each layer, we measured the fraction of significant neurons whose deviation area was above a heuristic threshold of $D=10$.

\section{Results}
\subsection{Dot whiteness experiment}
With increased dot whiteness, and subsequently increased pixel distance from the black dot image, we expected the representational dissimilarity $R$ to monotonically increase from $R=0$ at $\gamma=0$ to $R_{\text{max}}$ at $\gamma=1$ (Fig.~\ref{fig:fig1}c). The expected monotonic relation between dot whiteness and representational dissimilarity was evident in the ``No Lines'' control image (Fig.~\ref{fig:fig2}b), where the lines of the Scintillating Grid were removed to effectively eliminate illusion perception. The monotonic relation was also evident in most natural and synthetic images (Fig.~\ref{fig:fig2}c). Interestingly, we observed a significant deviation from the monotonic relation when increasing dot whiteness on images of the Scintillating Grid illusion (Fig.~\ref{fig:fig2}a). The representational dissimilarity increased in a monotonic fashion to an $R_{\text{max}}$ at $\gamma=0.55$ before decreasing and leveling off for higher whiteness levels ($\gamma>0.55$). This trend was noticeably different from the completely monotonic behavior observed in natural and synthetic images and grid illusion controls (Fig.~\ref{fig:fig2}d). The Scintillating Grid-specific deviation was observed for several different grid sizes, in images with different numbers of dots, and in translated grids (data not shown). This deviation from the expected monotonic behavior implicated a significant effect that was sensitive to dot whiteness, but independent of absolute differences in pixel values.

To quantify the magnitude of this illusion-like effect in VGG-19, we computed the deviation magnitude $d$ at each experimental interval. In the Scintillating Grid illusions, low deviation magnitudes were observed in early intervals ($\gamma < 0.55$) but increased after $\gamma=0.55$ (Fig.~\ref{fig:fig2}e). In comparison, the No Lines control produced minimal deviation magnitude throughout all dot whiteness intervals (Fig.~\ref{fig:fig2}e). As a result, the Scintillating Grid observed a much higher deviation area ($D=0.48$) as compared to the No Lines control ($D=0$).

\subsection{Natural and Synthetic Images and Grid Controls}
\begin{figure}
    \centering
    \includegraphics[width=\textwidth]{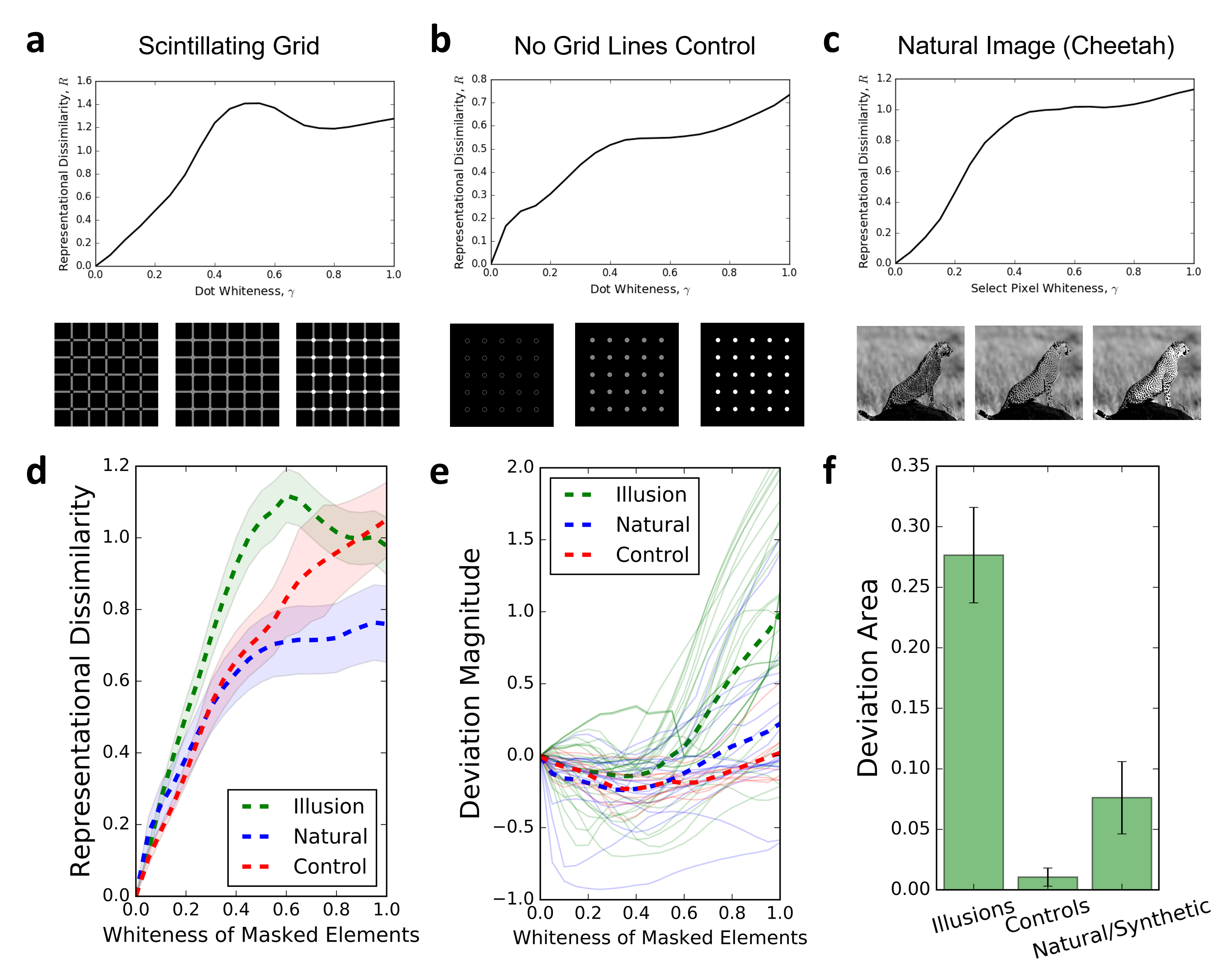}
    \caption{Significant deviations from the expected monotonic increase of representational dissimilarity $R$ with increasing whiteness $\gamma$ was selectively present in the Scintillating Grid. (a) The representational dissimilarity $R$, calculated as described in Fig.~\ref{fig:fig1}, for increasing dot whiteness $\gamma$ in the Scintillating Grid image. The measured $R$ deviated from the expected linear increase. (b) Removing lines from the Scintillating Grid eliminated human perception of the grid illusion, and recovered a clearly monotonic relation between $R$ and $\gamma$ which was approximately linear with increasing $\gamma$. (c) Representative example of a natural image control (cheetah). Although natural and synthetic images deviated from the expected linear relation, they exhibited significantly less deviation in VGG-19 $R$ than the Scintillating Grid and observed a monotonic relation between $R$ and $\gamma$. (d) Representational dissimilarity averaged across samples in three stimuli sets (30 illusions, 19 natural and synthetic images, 11 illusion controls). Shaded region represents standard error of the mean. (e) Deviation magnitude $d$ as a function of element whiteness for illusions, controls, and natural/synthetic images. Individual trajectories are shown in solid lines. (f) There was a significant difference between the mean deviation area $D$ of illusion and control stimuli sets. Error bars represent standard error of the mean $D$. Note that the magnitude of $R$ (y-axis scaling) is irrelevant in panel d because of differences in the number of masked pixels and presence of grid elements (see Supplementary for details).}
    \label{fig:fig2}
\end{figure}

We examined the robustness of the illusion-like effect in a diverse set of illusion variants ($n=11$) and control images ($n=19$). The majority of the control images observed strictly increasing representational dissimilarity $R$ when subjected to increasing masked region whiteness $\gamma$ (Fig.~\ref{fig:fig2}e). Of the images that showcased some deviation from the expected monotonic relation, none were as significant as those observed for the Scintillating Grid (independent t-test, $p=2.8\times 10^{-4}$ for mean $D$ values) (Fig.~\ref{fig:fig2}f). 

The mean deviation area for illusions ($D = 0.26 \pm 0.015$, Mean $\pm$ SEM) was significantly higher than that for illusion controls ($D = 0.011 \pm 0.075$) and natural and synthetic images ($D = 0.076 \pm 0.030$), which indicated a significant deviation in the VGG-19 model that was specific to grid illusions (Fig.~\ref{fig:fig2}f). Interestingly, the mean deviation area of the natural and synthetic images was higher than that of the illusion controls (Fig.~\ref{fig:fig2}f). Unlike illusion controls, which had contoured dots that prevented loss of color-derived boundaries under increasing whiteness (see Section 2.2), natural and synthetic images may be sensitive to these changes in color and contrast contours. This difference is a possible explanation for the inflated $D$ measurement observed in natural and synthetic images as compared to illusion controls (see Supplementary for more discussion).

Interestingly, a much less pronounced response was observed in ResNet \cite{he_deep_2016}, another deep convolutional neural network (see Supplementary). Unlike in VGG-19, the ResNet deviation area $D$ for Scintillating Grid variants was only significantly higher than the $D$ of natural and synthetic images.

\subsection{Number of white dots experiment}

In the previous experiment where the unit of change was dot whiteness, only the pixel difference between the two compared images was proportional to the unit of change. In that setup, we observed a significant competing effect outside of pixel differences, which resulted in a deviation from the expected monotonic increase in $R$ with increased dot whiteness (Fig.~\ref{fig:fig2}a). We postulate that this deviation represents VGG-19 ``perception" of a human-like illusion effect in the Scintillating Grid. To test this theory, we attempted to recover the expected behavior by using the number of white dots in the grid image as the unit of change in lieu of dot whiteness. In this setup, black-dotted Scintillating Grids with progressively increased numbers of white dots were compared to an all black dot grid (Fig.~\ref{fig:fig3}a). Like in the previous dot whiteness setup, the number of white dots is linearly proportional to the pixel difference between the two compared grids. However, unlike the previous setup, the number of white dots is additionally proportional to the magnitude of the perceived illusions for human observers since each white dot contributes equally to the illusion effect. Therefore, if the observed deviation is a model correlate of human illusion perception, then varying the number of white dots linearly would eliminate any deviation from a monotonic relation between $R$ and the number of white dots. 

As the number of white dots increased, the representational dissimilarity $R$ increased linearly as expected of a stimulus with competing pixel difference and illusion-like deviation effects (Fig.~\ref{fig:fig3}a). By constraining illusion-like effects to be linearly proportional to the units of change, we effectively eliminated the previously observed deviation. These results support human-like illusory perception as a contributing factor to the significant deviation observed in the VGG-19 representational dissimilarity for the Scintillating Grid in the previous dot whiteness setup where only pixel differences were held to be proportional to the units of change.

\subsection{Comparison to human vision}

\begin{figure}[H]
    \begin{center}        
    \includegraphics[width=\textwidth]{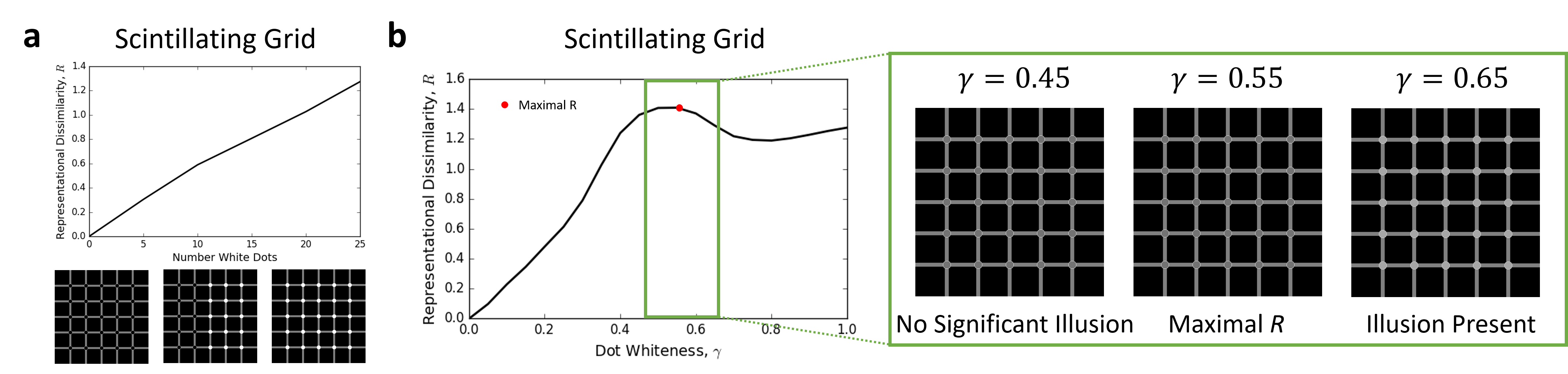}
	\end{center}
	\caption{(a) Increasing number of whites dots controlled for illusion-like deviations since the number of white dots is linearly proportional to the magnitude of the perceived illusions (for human observers) and to the pixel differences between the two compared images. Grids with incrementally greater numbers of white dots were compared to an all black dot grid. This setup recovered the expected linearly increasing representational dissimilarity with increased number of white dots. (b) Comparison to human perception. Visually, the transition from illusion-absent to illusion-present grid images is around $\gamma \approx 0.55$. This corresponded to the dot whiteness value with maximal VGG-19 representational dissimilarity $R$.}
	\label{fig:fig3}
\end{figure}

After characterizing the illusion-like response of VGG-19 to the Scintillating Grid, we compared it to previously reported measurements of human perception for the same illusion \cite{sun_characterizing_2019}. Results (Fig.~\ref{fig:fig2}a) showed that the dot whiteness interval with maximal VGG-19 representational dissimilarity roughly corresponded to the onset of significant illusion perception in the DNN model. That is, the standard Scintillating Grid stimulus induced a significant illusion-like response in VGG-19 for dot whiteness $\gamma>0.55$. Correspondingly, an earlier experiment with human observers ($n=8$) using the same visual stimuli, found that the dot whiteness critical point for illusion perception (defined as the $\gamma$ where half or less of the participants perceived the illusion) was around $\gamma_c = 0.60$ \cite{sun_characterizing_2019}. This similarity in perceptual ranges can be readily verified by visual inspection of Fig.~\ref{fig:fig3}b.

\subsection{Origin and propagation of illusion-like perturbation}

\begin{figure}[h!]
	\begin{center}        
    \includegraphics[width=\textwidth]{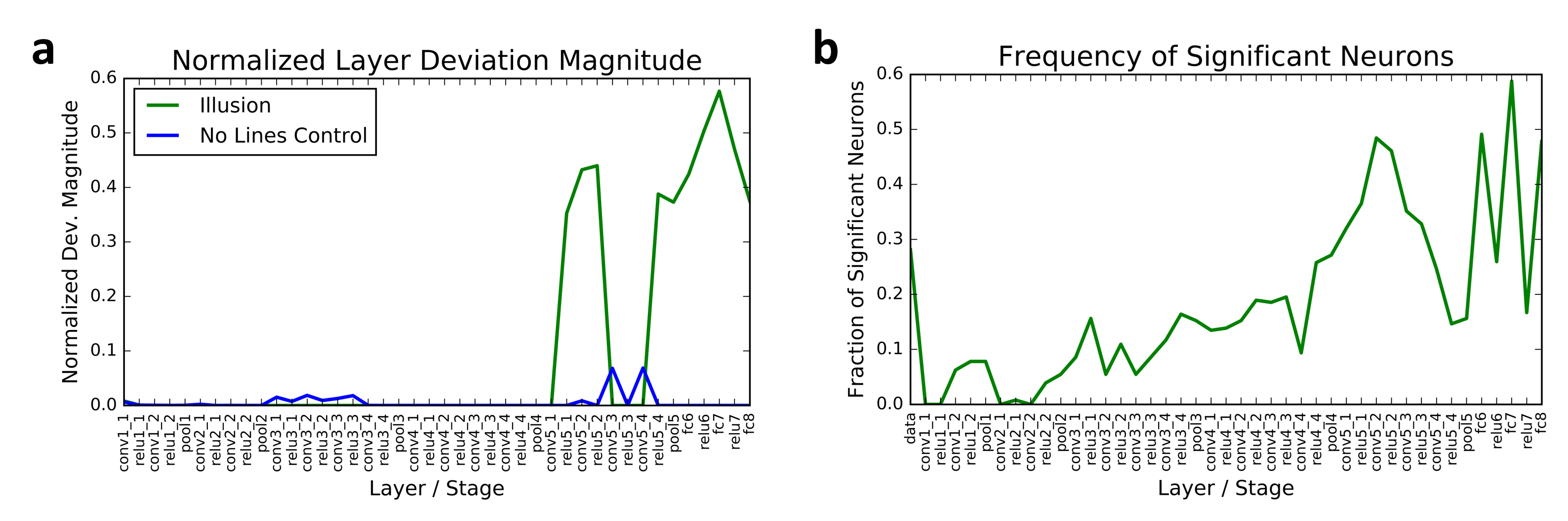}
	\end{center}
	\caption{(a) VGG-19 deviation area $D$ for each layer / computational stage of the DNN with respect to the Scintillating Grid and the No Lines control grid; (b) Fraction of significant neurons for each layer / computational stage determined using a heuristic threshold of $D=10$.}
	\label{fig:fig4}
	\end{figure}

We analyzed individual layer outputs to determine if any subset of layers was responsible for the origin or propagation of the illusion response. Results for the Scintillating Grid showed a sharp induction of an illusion-like effect in the deeper layers starting with \texttt{relu5\_1}, disappearing at \texttt{conv5\_3}, and then reappearing and persisting after \texttt{relu5\_4} up until the final layer \texttt{fc8} (Fig.~\ref{fig:fig4}a). To determine the number of significant neurons per layer, we applied the same method with a heuristic threshold of $D=10$ (Fig.~\ref{fig:fig4}b).

These results suggested that later processing stages were responsible for illusion-like response in VGG-19. However, most theories of grid illusion perception in human vision focus on earlier visual processing such as lateral inhibition in retinal ganglion cells or S1 simple cells in the primary visual cortex. Since a universal theory of grid illusion perception has not been established \cite{schiller_hermann_2005}, it may be worthwhile to consider the role of deeper visual processing in human perception of the Scintillating Grid.

We also used back-propagation to visualize regions of highest activation in VGG-19 for different image stimuli. These activation patterns were qualitatively different for the Scintillating Grid as compared to the No Lines control (see Supplementary for more details). Additionally, we applied principal component analysis to the layer $R$ values and observed that the first principal component was sufficient to resolve all whiteness levels in illusion controls and natural and synthetic images, while the first two principal components was necessary to fully resolve the different whiteness levels of the Scintillating Grid (see Supplementary).

\section{Discussion}

Here, we report that a human visual illusion, the Scintillating Grid, evoked a potential correlate of human illusory perception in VGG-19, a deep convolutional neural network. By measuring the representational dissimilarity $R$ between grid illusions of varying dot whiteness and a black dot grid illusion, we showed that the observed trends in $R$ deviated significantly from the expected monotonic behavior observed in natural and synthetic images and illusion control images. Varying the number of white dots in the grid illusion (and hence associating the magnitude of human-perceived illusion effect to pixel difference) recovered a linear relation between $R$ and the number of white grid dots. Overall, these results suggest that a strong nonlinear relation between $R$ and dot pixel whiteness $\gamma$ was present in Scintillating Grid variants for VGG-19 and that this effect was not simply the result of pixel differences. We propose the non-monotonic deviation in $R$ observed in VGG-19 as a model correlate of human illusory perception of the Scintillating Grid because:
\begin{enumerate}
    \item The Scintillating Grid produced the largest VGG-19 $R$ deviation from monotonic behavior among all natural and synthetic images and control grid images investigated in this study.
    \item When the number of white dots was increased rather than dot whiteness, the deviation was lost and the $R$ increased linearly with the number of white dots. This is consistent with a deviation that correlates to human perception of the Scintillating Grid illusion.
    \item The illusion-like effect was present for a dot whiteness range characterized by a critical threshold ($\gamma_c=0.55$) that was similar to that for human perception of the Scintillating Grid ($\gamma_c=0.60$) \cite{sun_characterizing_2019}.
\end{enumerate}
To our knowledge, these results are the first indication that a deep neural network may exhibit human-like representations of select visual illusions like the Scintillating Grid.

Several theories have been proposed to explain the neural mechanisms responsible for the perception of these grid illusions. The traditional theory posits that shallow retinal ganglion cell processes produce the observed center-surround effect \cite{spillmann_hermann_1994, wolfe_global_1984}. Retinal ganglion cells typically exhibit a center-surround receptive field, whereby the activity of some cells is increased by light falling on the excitatory center and decreased by light falling on the inhibitory surround (ON-center cells), or vice versa (OFF-center cells) \cite{enroth-cugell_contrast_1966}. The illusory perception of dark dots at the intersections of the Hermann Grid was therefore argued to manifest from more light falling on the inhibitory surround of ON-center cells at the intersections than at other areas flanked by single grid lines \cite{spillmann_hermann_1994, eagleman_visual_2001, amari_dynamics_1977}. Subsequent studies have indicated that the retinal ganglion cell theory is not sufficient to explain additional properties of grid illusions, which include the perception of the illusion under rotation and perturbation of the grid lines \cite{schiller_hermann_2005}. As such, it was proposed that additional downstream processing, such as in V1, is involved in mediating illusion perception for human observers.

In VGG-19, we observed large deviation magnitudes ($d$) and greater proportions of deviation-significant neurons in the deepest layers of VGG-19. Therefore, the illusion correlate is more likely to have originated from these deep layers rather than being propagated from the earlier processing. Although a direct correlation between the architectures of human and computer vision is difficult, early DNN layers may be comparable to the human opponent-color and frequency-selective representations in retinal ganglion cells and in V1 neurons, while deeper DNN layers might compare with higher areas in the human visual system, such as V4 or IT. Consideration of higher level processing in human perception of the Scintillating Grid and its variants may potentially yield new insight into the underlying mechanisms of visual illusion perception. Additionally, as human perception of the Scintillating Grid is restricted to the periphery, VGG-19 could potentially serve as a model correlate of human peripheral vision. 

The main contribution of this work is to outline a novel model correlate of human illusion perception
in a DNN: non-monotonicity in representational distance. It is important to note that the evidence
provided here is still preliminary. First and foremost, it is not clear what is the scientific
reason that the non-monotonicity is a model correlate for perception of the Scintillating Grid
illusion (beyond the intuitive idea that in the presence of illusion the white dots are more ”similar” to
the black dots). This is a broad question, which may be addressed by investigating non-monotonicity
in the DNN, or illusion perception in humans. Second, further validation is required to strengthen
the current experiments. This includes replication using a larger set of illusion variants, illusion controls,
and natural and synthetic images, the use of other deep learning models, other representational
distance metrics (e.g., cosine distance, although it would be insensitive to differences in absolute
magnitudes between the two representations), and other masking schemes in the control images
(e.g. masking regions that correspond to the dots of the Scintillating Grid, as opposed to masking
based on a luminance intensity threshold which was used here). Additionally, an adversarial setup could be leveraged to search latent space for images that exhibit a non-monotonic relation between representational dissimilarity and mask whiteness, which may lead to the discovery of novel grid-like visual illusions. Since correlations have been reported between VGG-19 and human categorizations of competing images, it may be of interest to examine the role of perceptual dominance in the perception of the Scintillating Grid in both human and computer vision models \cite{gruber_perceptual_2018}.

In conclusion, the work suggests a novel area of overlap between human and computer vision, which we hope will motivate further investigation of visual illusions using computer vision, and vice versa.

\section*{Acknowledgements}
We would like to thank the Dr. Bessie F. Lawrence International Summer Science Institute (ISSI) and Prof. Dov Sagi for the opportunity to initiate this collaboration.

\bibliographystyle{acm} 
\bibliography{VisualIllusions}

\raggedbottom
\pagebreak

\section{Supplementary Material}

\subsection{Stimuli sets}

\begin{figure}[H]
    \centering
    \includegraphics[width=\textwidth]{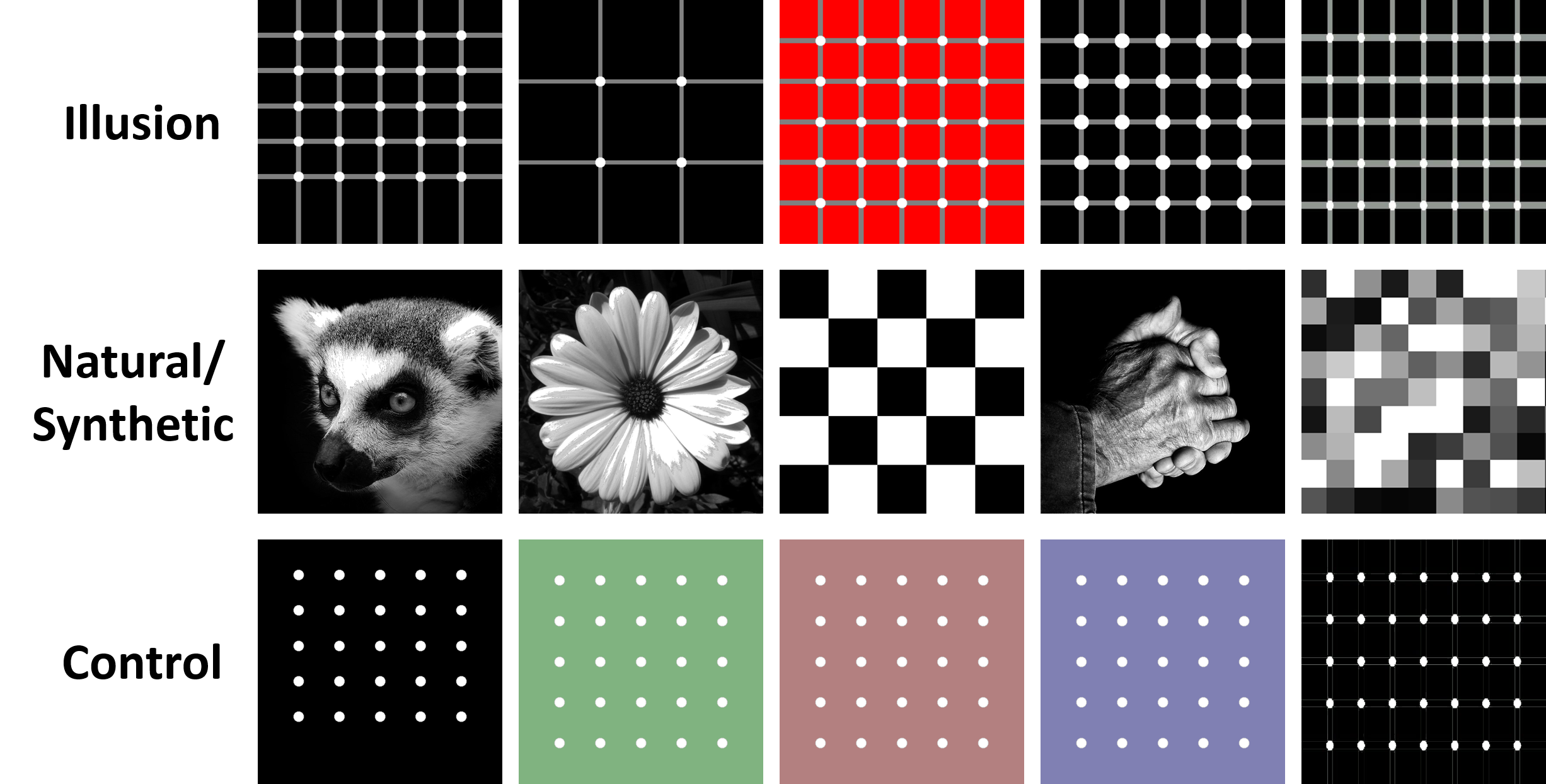}
    \caption{Representative images from each of the three stimuli sets: illusion variants, natural and synthetic images, and illusion control images.}
    \label{fig:stimuli}
\end{figure}

We used three image stimuli sets in the masked element and dot whiteness experiments. These included 30 illusion variants, 19 natural and synthetic images, and 11 illusion control images. In Figure \ref{fig:stimuli}, five representative images from each of the three sets are depicted for reference. All image sets are available on the public Github repository: \url{https://github.com/sunericd/dnn-illusion}.

\subsection{Whole-network representational dissimilarity pattern}
\begin{figure}[h!]
\begin{center}
\includegraphics[width=1.0\textwidth]{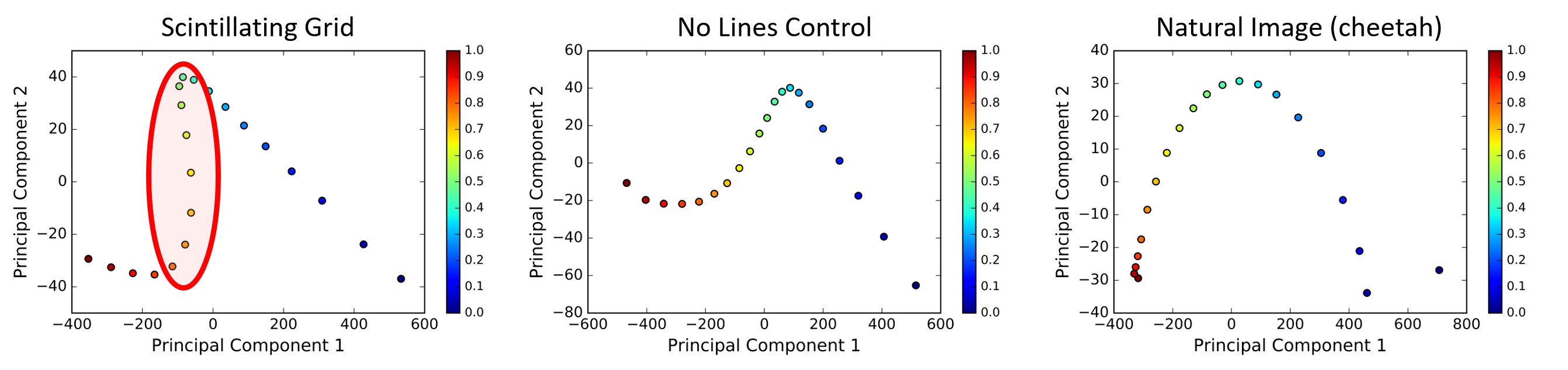}
\caption{Principal component analysis (PCA) of VGG-19 representational dissimilarity $R$ vectors of all layers. Shown are the first two principal components for the Scintillating Grid, No Lines control, and an example natural image (cheetah). Colors correspond to the dot whiteness or masked element whiteness level $\gamma$. The first principal component is insufficient for discriminating all dot whiteness intervals in the Scintillating Grid (see red shaded region) but is sufficient to discriminate intervals in the other control stimuli.}
\label{supp:pca}
\end{center}
\end{figure}

Our analyses used the outputs of \texttt{fc8}, which was the final fully convolutional layer of VGG-19 and arguably the closest analog to perception in human vision since it is the stage just prior to classification. To understand the dissimilarity patterns across all layers, principal component analysis (PCA) was applied to the representational dissimilarity $R$ vectors of each layer/stage in VGG-19 (as compared to $D$ and $d$ values in Fig.~\ref{fig:fig4}). In illusion controls and natural and synthetic images, the first principal component was sufficient to discriminate the different dot whiteness illusion variants (i.e.~the values corresponding to different dot whiteness levels were separated when projected onto the first principal component as is shown in Fig.~\ref{supp:pca}bc). On the other hand, the values corresponding to dot whiteness in the Scintillating Grid were significantly crowded along the first principal component and the second principal component was necessary to fully resolve individual whiteness intervals (see Fig.~\ref{supp:pca}a). The median whiteness level in the crowded region was $\gamma=0.60$, which was close to the previously observed $\gamma_\text{max}=0.55$ in \texttt{fc8}. This implies that the sources of variance captured by the first principal component are primarily responsible for illusion perception since the component fails to fully discriminate the dot whiteness intervals corresponding to illusion perception. It will be of interest to characterize the correlates of the first principal component in human vision and its possible contribution to illusion perception.

\subsection{VGG-19 Visualization}

To develop a visual understanding of VGG-19 ``perception" of the Scintillating Grid, we utilized methods for visualizing DNN activation. One approach for visualizing regions of greatest activation for a given neuron is to use a backward pass for the neuron activation after the forward pass by the network. The gradient of the activation is then associated with the activation levels and was visualized as such. This can also be achieved with a gradient of the class score with respect to the input image \cite{simonyan_deep_2013}, which tends to offer a better visualization of the entire network than vanilla back-propagation. We adopted a variant of this approach and used gradient visualization with back-propagation through guidance from ''deconvnet" visualization \cite{springenberg_striving_2014}. The implementation was adapted from the ``pytorch-cnn-visualizations" project for a pre-trained VGG-19 model.

\begin{figure}[H]
		\begin{center}        
        \includegraphics[width=1.0\textwidth]{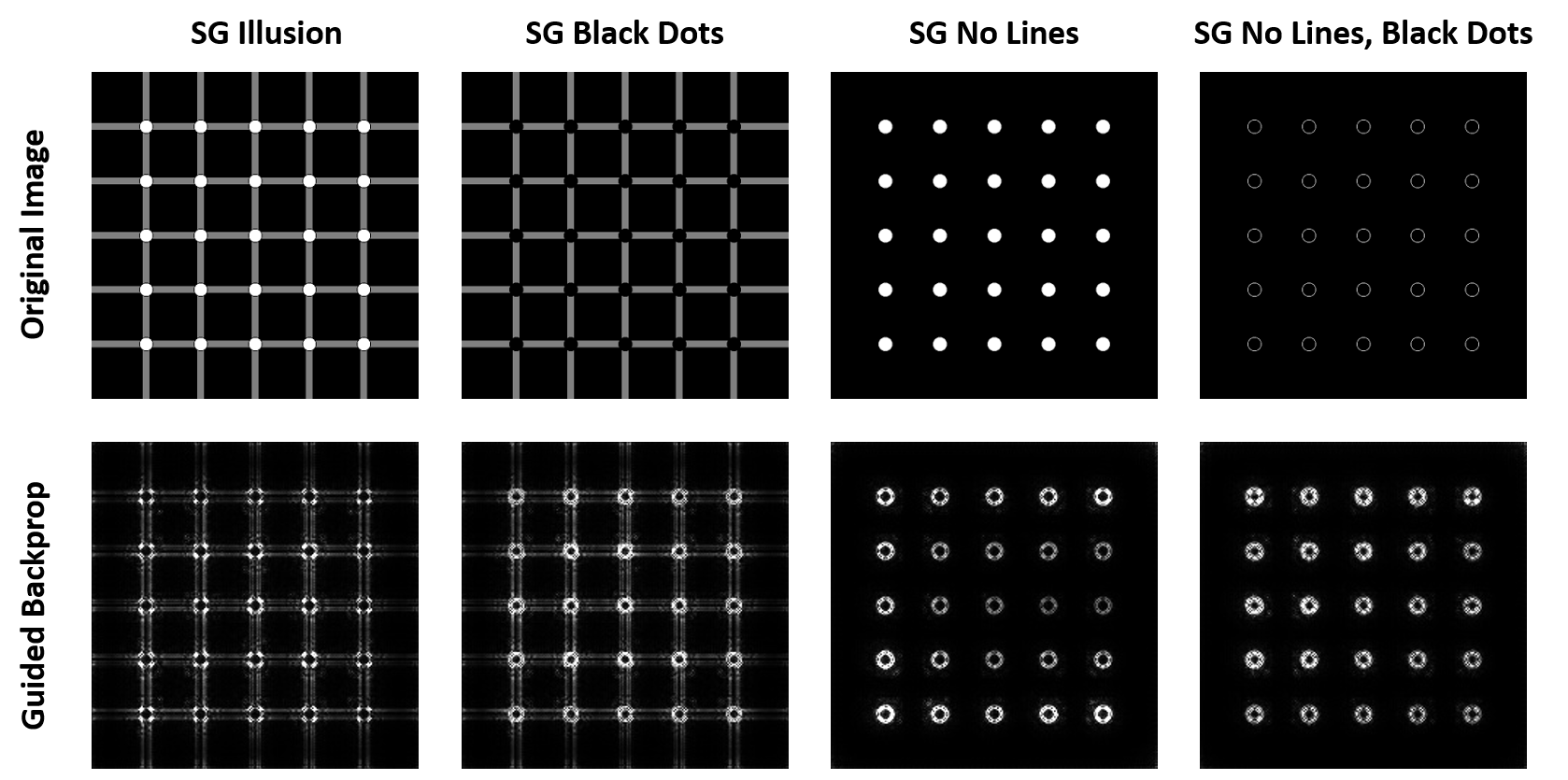}
		\end{center}
		\caption{Gradient visualization of the ImageNet VGG-19 activation with respect to the Scintillating Grid and its non-illusory variants using vanilla back-propagation and guided back-propagation.}
		\label{supp:visualization}
	\end{figure}

We visualized the activation of the VGG-19 with respect to the Scintillating Grid and three control variants that exhibited no illusion effect: the Scintillating Grid with black dots, the Scintillating Grid with no lines, and the Scintillating Grid with black dots and no lines. This was achieved using guided back-propagation. Guided back-propagation \cite{springenberg_striving_2014} was performed on the class differentials in the final output layer of VGG-19. Applying guided back-propagation on the Scintillating Grid illusion revealed a disjointed, quad-like arrangement of high activation patches around the border of each dot. This pattern was not observed in the corresponding patterns of activation of the three non-illusion images, which mostly consisted of continuous, circular boundaries of activation around the dots (Fig.~\ref{supp:visualization}).

\subsection{ResNet shows no illusion-like response}

\begin{figure}[H]
    \centering
    \includegraphics[width=\textwidth]{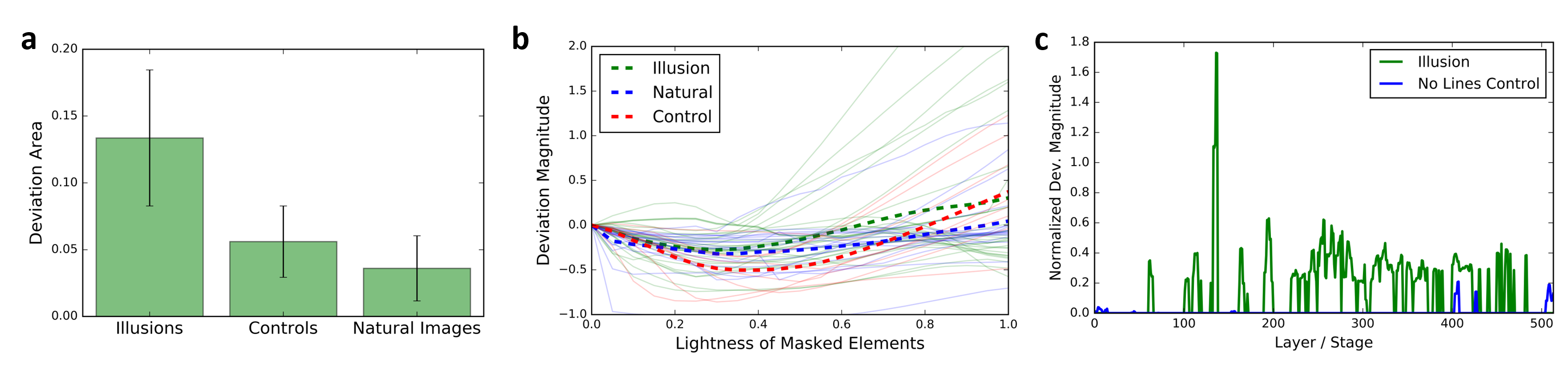}
    \caption{ResNet exhibits no significant illusion-like deviation. (a) Deviation area for illusion variants, natural and synthetic images, and illusion controls with standard error of mean. (b) Deviation magnitudes for different element whiteness intervals for illusion variants, natural and synthetic images, and illusion controls. (c) Deviation area across different ResNet stages/layers for the Scintillating Grid and the No Lines control.}
    \label{fig:resnet}
\end{figure}

We investigated the deep convolutional ResNet-152 network \cite{he_deep_2016} for illusion-like responses. For the same 30 illusion stimuli set and 30 control (19 natural, 11 illusion control) stimuli set, there was a less pronounced illusion-like response as characterized by the deviation areas between illusions ($D = 0.0830 \pm 0.0286$) and controls ($D = 0.0360 \pm 0.0244$ for natural, $D = 0.05603 \pm 0.0267$ for illusion controls) (Fig.~\ref{fig:resnet}a). The deviation magnitude trajectories corroborated this trend (Fig.~\ref{fig:resnet}b).

This difference as compared to the responses of VGG-19 suggests that the illusion-like effect is sensitive to network architecture--VGG-19 consists of 19 layers while the implementation of ResNet is much deeper with 152 layers \cite{he_deep_2016}. Comprehensive analyses of other deep neural network models may further inform the properties which influence the susceptibility of a DNN to perception of the Scintillating Grid.

\subsection{Comment on the lack of deviation magnitude comparability between stimuli sets}

The magnitude of representational dissimilarity $R$, and therefore $D$ and $d$, were not comparable between the different stimuli sets due to intrinsic differences in the size and placement of the masked regions, and due to differences in the backgrounds (non-masked) regions. These differences were not trivially amenable to normalization methods due to the complexity of factors involved (e.g.~different baseline levels of white pixels, different average pixel value, etc). Furthermore, since the competing illusion-like effect is additive, normalization may artificially inflate or deflate the illusion scoring statistics. Therefore, we only considered the monotonicity of the relation of $R$ with respect to dot whiteness.

\end{document}